\documentclass{./styles/svproc}
\usepackage{url}

\usepackage{graphicx}
\usepackage{amsmath}
\usepackage{amssymb}
\usepackage{utfsym}
\usepackage{multirow}
\usepackage{booktabs}
\usepackage{caption}
\usepackage{subcaption}
\usepackage[utf8]{inputenc}
\usepackage{diagbox}
\usepackage{hyperref}
\hypersetup{
    colorlinks=true,
    linkcolor=blue,
    filecolor=magenta,      
    urlcolor=red,
    pdftitle={Overleaf Example},
    pdfpagemode=FullScreen,
    }

\begin{document}
\mainmatter              % start of a contribution
\title{Label-free Motion-Conditioned Diffusion Model for Cardiac Ultrasound Synthesis}
\titlerunning{Label-Free Motion-Conditioned Diffusion Model}  % abbreviated title (for running head)
\author{Zhe Li \inst{1}\(^\ast\) \and Hadrien Reynaud\inst{2} \and
Johanna P Müller\inst{1} \and Bernhard Kainz\inst{1,2}}
\authorrunning{Zhe Li et al.} % abbreviated author list (for running head)
% \tocauthor{Ivar Ekeland, Roger Temam, Jeffrey Dean, David Grove}
%
% \institute{Princeton University, Princeton NJ 08544, USA,\\
% \email{I.Ekeland@princeton.edu},\\ WWW home page:
% \texttt{http://users/\homedir iekeland/web/welcome.html}
% \and
% Universit\'{e} de Paris-Sud,
% Laboratoire d'Analyse Num\'{e}rique, B\^{a}timent 425,\\
% F-91405 Orsay Cedex, France}
\institute{Department AIBE, FAU Erlangen-Nürnberg, Erlangen, Germany, \\
\email{zhe.li@fau.de},\\
\and
Department of Computing, Imperial College London, London, UK}

\maketitle              % typeset the title of the contribution

\begin{abstract}
Ultrasound echocardiography is essential for the non-invasive, real-time assessment of cardiac function, but the scarcity of labelled data, driven by privacy restrictions and the complexity of expert annotation, remains a major obstacle for deep learning methods. We propose the Motion Conditioned Diffusion Model (MCDM), a label-free latent diffusion framework that synthesises realistic echocardiography videos conditioned on self-supervised motion features. To extract these features, we design the Motion and Appearance Feature Extractor (MAFE), which disentangles motion and appearance representations from videos. Feature learning is further enhanced by two auxiliary objectives: a re-identification loss guided by pseudo appearance features and an optical flow loss guided by pseudo flow fields. Evaluated on the EchoNet-Dynamic dataset, MCDM achieves competitive video generation performance, producing temporally coherent and clinically realistic sequences without reliance on manual labels. These results demonstrate the potential of self-supervised conditioning for scalable echocardiography synthesis. Our code is available at \url{https://github.com/ZheLi2020/LabelfreeMCDM}.
\end{abstract}

\section{Introduction}
\label{sec:intro}

Cardiac conditions such as abnormal Ejection Fraction (EF) and early valvular disease often present with subtle imaging patterns in ultrasound~\cite{bloom2017heart,omar2016advances}, placing a high demand on expert interpretation and consistency in image acquisition. Supervised learning methods are particularly constrained in this context, as they rely on large, diverse, and expertly annotated datasets that are rarely attainable. In echocardiography, these difficulties are further compounded by inter-observer inconsistency, cross-platform differences, and data-sharing constraints imposed by institutional privacy policies, which limit the assembly of comprehensive training cohorts. Overcoming these barriers is essential for advancing scalable, data-driven methods in tasks such as video classification, regression, and segmentation.

To mitigate data scarcity, recent works have explored generative models such as GANs~\cite{tomar2021content,reynaud2022d} and diffusion models~\cite{nguyen2024training,reynaud2024echonet,stojanovski2023echo} for synthetic data generation. These methods are promising but typically rely on clinical labels (\emph{e.g.}, ejection fraction or segmentation maps), which are costly to obtain and limit scalability. In contrast, self-supervised learning offers a powerful alternative by exploiting the inherent structure of unlabelled videos. This is particularly compelling in echocardiography, where vast amounts of unannotated data are already available but remain underutilised, providing a direct pathway to scalable and label-free generative modelling.

In this work, we propose the \emph{Motion Conditioned Diffusion Model} (MCDM), a self-supervised framework for ultrasound video synthesis that removes the need for clinical labels. MCDM trains a latent video diffusion model conditioned on motion features rather than EF scores, leveraging temporal dynamics that are more informative than the highly similar static appearance across echocardiography videos.
To extract these features, we introduce the Motion-Appearance Feature Extractor (MAFE), which disentangles motion and appearance representations, inspired by architectures developed for video frame interpolation~\cite{zhang2023extracting}. As a preprocessing step, a ReID model~\cite{dar2024unconditional} generates appearance embeddings, and UnSAMFlow~\cite{yuan2024unsamflow}, an unsupervised optical flow model, produces flow fields. These precomputed representations act as pseudo ground truth to supervise MAFE, leading to robust appearance representations and reliable motion feature learning. To the best of our knowledge, this is the first work to apply self-supervised motion conditioning to a latent diffusion model for ultrasound echocardiography.

\noindent\textbf{Contributions: }
(1) We propose the first self-supervised Motion Conditioned Diffusion Model (MCDM) for medical video generation, which conditions on motion features to synthesise realistic echocardiography without annotated labels.
(2) We design a novel Motion-Appearance Feature Extractor (MAFE) that explicitly disentangles motion and appearance features, supervised by pseudo appearance embeddings and optical flow to deliver robust feature learning and improve temporal coherence.
(4) We evaluate MCDM on the EchoNet-Dynamic~\cite{ouyang2020video-based} dataset, showing that it achieves high-quality video generation and offers strong potential for scalable, annotation-free clinical applications.

\noindent\textbf{Related work: } 
Most unconditional video generation models on natural datasets rely on either U-Net~\cite{mei2023vidm,lapid2023gd,yu2023video,kim2024hybrid,wang2024leo} or Transformer~\cite{ge2022long,ma2024latte} backbones, synthesising images and extending them to videos using real or generated optical flow. 
In echocardiography, diffusion-based methods are all conditional. Reynaud et al.~\cite{reynaud2023feature,reynaud2024echonet} proposed a cascaded diffusion framework conditioned on a single image together with the EF value. Chen et al.~\cite{chen2024ultrasound} introduced the Latent Dynamic Diffusion Model (LDDM), which first captures temporal dynamics in a latent space and then decodes them into videos conditioned on static images. Nguyen et al.~\cite{nguyen2024training} generated echocardiogram videos from a single end-diastolic segmentation map using a 3D U-Net with temporal attention.
In contrast, our method introduces a self-supervised, label-free conditional diffusion model that leverages motion features extracted directly from unlabelled echocardiography videos for video synthesis.
For motion representation, Zhang et al.~\cite{zhang2023extracting} designed an inter-frame attention module to disentangle motion and appearance features, while Guo et al.~\cite{guo2024generalizable} estimated bilateral optical flow between adjacent frames using an adaptive coordinate-based network.
In addition, we generate two type of pseudo ground truth to enhance feature learning.

\section{Method}
\label{sec:method}
We propose a self-supervised framework for ultrasound video synthesis that removes reliance on clinical annotations. 
The framework consists of three main components. First, a feature extractor is trained to disentangle motion and appearance representations (Section~\ref{sec:featureextractor}). Second, its training is guided by two auxiliary objectives: a re-identification loss using pseudo appearance features and a flow-based loss using pseudo optical flow (Section~\ref{sec:reidloss}). Finally, a latent diffusion model, conditioned on the precomputed motion features, is trained to synthesise echocardiography videos (Section~\ref{sec:uncondtraining}).

\begin{figure}[tb]
\includegraphics[width=\textwidth]{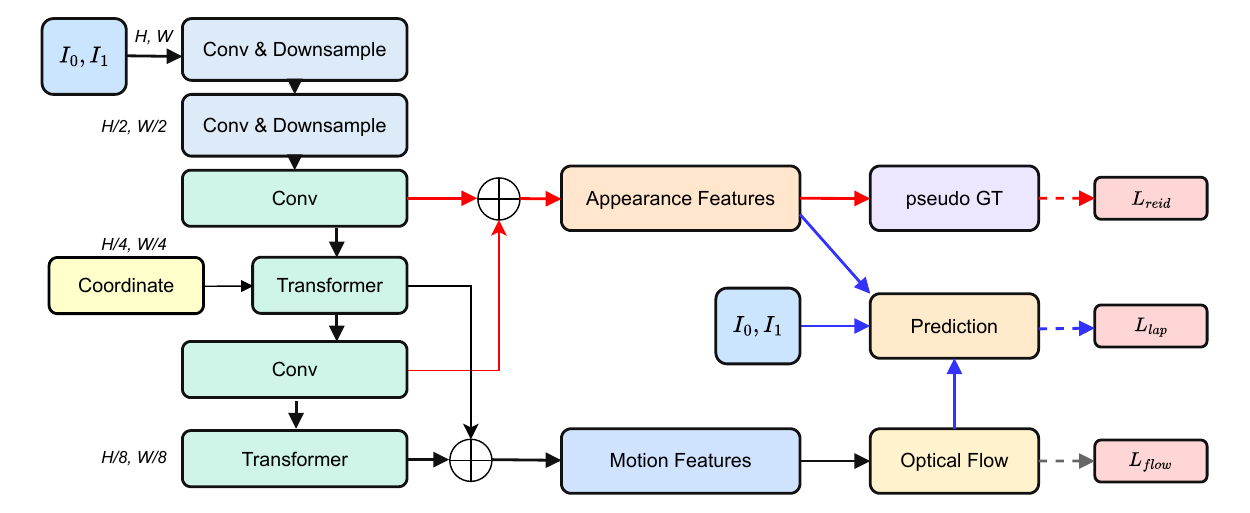}
\caption{Overview of the proposed Motion-Appearance Feature Extractor (MAFE). Given two input frames \(I_0\) and \(I_1\), the network disentangles appearance and motion features. A re-identification loss guided by pseudo appearance features improves appearance learning, while a flow loss guided by pseudo optical flow refines motion representations.} 
\label{fig:motionextractor}
\end{figure}

\subsection{Motion-appearance feature extractor}
\label{sec:featureextractor}
Fig.~\ref{fig:motionextractor} illustrates the architecture of our proposed Motion-Appearance Feature Extractor (MAFE), inspired by the EMA-VFI framework~\cite{zhang2023extracting} originally developed for video frame interpolation. EMA-VFI explicitly disentangles motion and appearance representations, unlike most existing networks that extract them jointly or in parallel. This property makes it well suited for echocardiography, where high visual similarity reduces the usefulness of appearance features and subtle temporal dynamics carry greater diagnostic value. 

Building on this principle, we design MAFE specifically for ultrasound. 
The network takes two input frames \(I_0\) and \(I_1\), typically chosen as End-Diastolic (ED) and End-Systolic (ES) to maximise observable motion. These frames can be identified without expert clinical annotation, while the framework remains flexible to other frame pairs when ED/ES landmarks are unavailable. To better preserve structural details in low-resolution ultrasound data and reduce overfitting on relatively small datasets, we design a shallower architecture by removing one convolutional and downsampling block compared to EMA-VFI. 
Appearance features are first computed by the convolutional blocks, and directional motion features are then derived using these appearance features and spatial coordinate matrices. Specifically, \(F_{0 \rightarrow 1}\) encodes motion from  \(I_0\) to \(I_1\), and \(F_{1 \rightarrow 0}\) captures the inverse. These are concatenated to form a bidirectional motion features. Finally, the motion and appearance features, together with the original frames, are passed through a lightweight convolutional module that estimates intermediate motion flow and predicts the middle frame $\tilde{I}$  between \(I_0\) and \(I_1\).

%---------------------------------------------------------------------

\subsection{Pseudo ground truth supervision}
\label{sec:reidloss}

To further enhance feature learning, we introduce two auxiliary objectives: a re-identification loss guided by pseudo appearance features to improve appearance learning, and a flow loss guided by pseudo optical flow to refine motion features.

\noindent\textbf{Pseudo appearance features. } To guide appearance feature learning, we train a re-identification (ReID) model~\cite{dar2024unconditional} using contrastive learning to generate frame-level embeddings. Image pairs are labelled as positive if they are from the same video, and negative if they are from different videos. The ReID model learns to minimise the distance between embeddings of positive pairs while maximising the distance for negative pairs. For compatibility with MAFE, the embedding dimension is set to match that of the appearance features. Once trained, the ReID model produces pseudo ground truth appearance embeddings for all input frames. These embeddings are then used to supervise MAFE through a mean squared error (MSE) loss, denoted as $\mathcal{L}_{reid}$.

\noindent\textbf{Pseudo optical flow. } To refine motion features, we employ the unsupervised optical flow model UnSAMFlow~\cite{yuan2024unsamflow}, which leverages segmentation masks generated by SAM~\cite{kirillov2023segment} to guide learning without ground-truth flow. For ultrasound videos, we train UnSAMFlow with the assistance of MedSAM~\cite{ma2024segment}, a segmentation model specifically developed for medical imaging, which in our case provides accurate cardiac chamber masks. After training, optical flow fields are generated and stored, serving as pseudo ground truth for MAFE. This supervision is applied through an MSE loss, denoted as $\mathcal{L}_{flow}$.

\noindent\textbf{Training objectives. } To ensure the quality of frame reconstruction, we adopt a Laplacian pyramids loss~\cite{niklaus2020softmax,zhang2023extracting}, defined as the L1 distance between the Laplacian pyramids of the predicted frame $\tilde{I}$ and the target frame $I_{GT}$:

\begin{equation}\label{eq:laploss}
    \mathcal{L}_{\text{Laplacian}}(\tilde{I}, I_{GT}) = \sum_{l=0}^{L} \left\Vert P(\tilde{I}^l) - P(I_{GT}^l) \right\Vert_1,\ P(I) = I - \text{up}\left(\text{down}(\psi(I))\right)
\end{equation}

where $l$ denotes the pyramid level, $\psi$ is a Gaussian filter, and $P$ represents the pyramid operation.
The final loss combines three terms as:

\begin{equation}\label{eq:totalloss}
    \mathcal{L} = \mathcal{L}_{\text{Laplacian}}(\tilde{I}, I) + \lambda_1\mathcal{L}_{reid} + \lambda_2\mathcal{L}_{flow}
\end{equation}

where $\lambda_1$ and $\lambda_2$ are hyperparameters balancing the contributions of the auxiliary objectives.

\subsection{Label-Free Diffusion Training}
\label{sec:uncondtraining}

After training MAFE, we generate motion features for each ultrasound video. These features are reshaped into 1D vectors and stored in advance for conditioning the diffusion model.
Our diffusion framework builds upon EchoSyn~\cite{reynaud2024echonet}, which consists of three components: a Variational Auto-Encoder (VAE), a Latent Image Diffusion Model (LIDM), and a Latent Video Diffusion Model (LVDM). We focus on training the LVDM, implemented as a Spatio-Temporal U-Net~\cite{blattmann2023stable} with four residual blocks.

Unlike EchoSyn, which conditions the diffusion process on expert-annotated EF scores, we propose the Motion Conditioned Diffusion Model (MCDM), which is conditioned directly on the precomputed motion features. MCDM is trained with an MSE loss in the latent space, enabling the synthesis of realistic and temporally coherent echocardiography videos without any manual annotations. This framework provides a scalable, label-free solution for both clinical and research applications.

\section{Experiments}

\subsection{Datasets and metrics} 
We evaluate our framework on the EchoNet-Dynamic dataset~\cite{ouyang2020video-based}, which contains $10,030$ ultrasound videos. 
This dataset is divided into $7,465$ for training, $1,288$ for validation, and $1,277$ for testing samples. The video resolution is $112 \times 112$, with sequence lengths ranging from $28$ to $1002$ frames, reflecting real-world variability.
The feature extractor is assessed using standard image quality metrics, namely Peak Signal-to-Noise Ratio (PSNR) and Structural Similarity Index Measure (SSIM). The diffusion model is evaluated using Fréchet Inception Distance (FID), Inception Score (IS), and Fréchet Video Distance at \(16\) frames (FVD$_{16}$) and \(128\) frames (FVD$_{128}$). 
Lower FID/FVD values and higher IS scores indicate better generative quality.

\subsection{Implementation details} 
For training the Motion-Appearance Feature Extractor (MAFE) on the EchoNet-Dynamic dataset, we use a batch size of $8$ and an initial learning rate of $2 \times 10^{-4}$, with cosine learning rate scheduling and $2{,}000$ warm-up steps. The feature extractor is configured with embedding channels $[64, 128, 256, 512]$ across four layers. The hyperparameters are set to $\lambda_1 = 1$ and $\lambda_2 = 0.01$, balancing the re-identification and flow loss terms.
During inference, bidirectional motion features are generated with spatial dimensions $[2, 256, 28, 28]$ and $[2, 512, 14, 14]$. The last two dimensions are averaged, and the resulting features are flattened and concatenated to produce a final motion feature vector of size $[1, 3072]$.
The diffusion model is trained from scratch for \(7\) days on $4 \times$ NVIDIA A100 GPUs (80GB), with a global batch size of \(64\) and a learning rate of $1 \times 10^{-4}$. 
All other settings follow the baseline configuration in EchoNet-Synthetic~\cite{reynaud2024echonet}.

\begin{table}[tb]
  \caption{Comparison of motion extractor performance using PSNR and SSIM.}
  \label{tab:featureextractor}
  \begin{center}
  \begin{tabular*}{0.8\textwidth}{@{\extracolsep{\fill}} l c c c c @{}}
    \toprule
     & Baseline & ReID & Flow & MAFE (ours) \\
    \midrule
    PSNR \(\uparrow\) & 23.17$_{\pm 1.76}$ & 23.24$_{\pm 1.76}$ & 23.24$_{\pm 1.74}$ & \textbf{23.31$_{\pm 1.75}$} \\
    SSIM \(\uparrow\) & 0.7528$_{\pm 0.05}$ & 0.7543$_{\pm 0.05}$ & 0.7520$_{\pm 0.05}$ & \textbf{0.7543$_{\pm 0.05}$} \\
    \bottomrule
  \end{tabular*}
  \end{center}
  \vspace{-3mm}
\end{table}

\subsection{Results and discussion} 

\noindent\textbf{Evaluation MAFE. } Table~\ref{tab:featureextractor} reports the performance of our motion feature extractor MAEF compared with several baselines. 
\texttt{Baseline} denotes the original EMA-VFI model~\cite{zhang2023extracting} without modifications. 
\texttt{ReID} refers to the model with structural adjustments trained using an additional re-identification loss with MSE supervision. \texttt{Flow} incorporates structural adjustments together with an optical flow loss. 
\texttt{MAFE (ours)} includes both modifications, combining the re-identification and flow losses with the adjusted architecture.
The results demonstrate a consistent improvement in performance as each modification is introduced. 
This confirms that both architectural tuning and pseudo ground truth guided auxiliary losses contribute positively to the training process, leading to more effective motion feature learning.

\begin{table}[tb]\setlength{\tabcolsep}{3pt}

\caption{Comparison with baselines and state-of-the-art diffusion models.
Methods marked with $\checkmark$ are conditioned on manual EF labels. In contrast, our proposed MCDM (marked with $\times$) is \textbf{label-free}, relying on motion features extracted in a self-supervised manner.}
\label{tab:diffusionsota}
\begin{center}
\resizebox{0.98\textwidth}{!}{%
\begin{tabular}{clclccccccc}
\toprule
Frames & Conditions & Labels & Method & FID$\downarrow$ & FVD$_{16}\downarrow$ & FVD$_{128}\downarrow$ & IS$_{\pm \text{std}}\uparrow$ \\
\hline
 & EF & \(\checkmark\) & EchoDiff.~\cite{reynaud2023feature} & 24.0 & 228 & - & 2.59$_{\pm 0.06}$ \\
& EF &\(\checkmark\) & EchoSyn~\cite{reynaud2024echonet} & 17.4 & 71.4 & 168.3 & 2.31$_{\pm 0.08}$\\
\hline
64 & EF+motion &\(\checkmark\) & baseline & 25.9 & 131.1 & - & 2.45$_{\pm 0.07}$\\
128 & EF+motion &\(\checkmark\) & baseline & 32.3 & 136.0 & 360.8 &  2.31$_{\pm 0.05}$ \\
\hline
\multirow{4}{*}{64}
& motion+ap & \(\times\) & baseline & 47.9 & 249.3 & -- & 2.17$_{\pm 0.05}$ \\
 & motion & \(\times\) & baseline & 48.8 & 244.2 & -- & 2.15$_{\pm 0.06}$ \\
& motion & \(\times\) & MCDM (ReID) & \textbf{46.0} & 236.1 & -- & 2.13$_{\pm 0.05}$ \\
& motion & \(\times\) & MCDM (ReID+Flow) & 47.1 & \textbf{225.1} & -- & \textbf{2.19$_{\pm 0.06}$} \\

\hline
\multirow{4}{*}{128} & motion+ap & \(\times\) & baseline & \textbf{60.8} & 291.6 & 816.8 & \textbf{1.97$_{\pm 0.05}$} \\
 & motion & \(\times\) & baseline & 67.6 & 286.7 & 566.6 & 1.93$_{\pm 0.04}$ \\
& motion & \(\times\) & MCDM (ReID) & 61.4 & 283.1 & 622.9 & 1.95$_{\pm 0.08}$ \\
& motion 64 & \(\times\) & MCDM (Flow) & 56.8 & 256.0 & 636.2 & 1.98$_{\pm 0.04}$ \\

& motion & \(\times\) & MCDM (ReID+Flow) & 62.5 & \textbf{262.8} & \textbf{553.2} & 1.95$_{\pm 0.06}$ \\
\bottomrule
\end{tabular}
}
\end{center}
\vspace{-3mm}
\end{table}

\noindent\textbf{Evaluation MCDM. } Table~\ref{tab:diffusionsota} reports the performance of our self-supervised Motion Conditioned Diffusion Model (MCDM). Since no prior work has investigated unconditional or self-supervised ultrasound video generation, we construct several baselines for comparison. 
The \texttt{EF + motion} baseline is initialised from EF-conditioned checkpoints and further trained with both EF and motion features. 
The \texttt{motion} baseline is trained from scratch using only motion features extracted from the original EMA-VFI. 
The \texttt{motion + ap} baseline incorporates both motion and appearance features as conditioning signals. Adding appearance features slightly improves FID (\(47.9\) vs. \(48.8\)) for 64-frame synthetic videos. However, it degrades temporal coherence, as reflected in poorer FVD$_{128}$ scores (\(816.8\) vs. \(566.6\)) for 128-frame videos, indicating that appearance conditioning reduces temporal realism.
The \texttt{motion + MCDM (ReID)} setting is conditioned on motion features extracted by MAFE trained with the re-identification loss, improving FID (\(61.4\) vs. \(67.6\)), but results in a worse FVD$_{128}$ (\(622.9\) vs. \(566.6\)).
Our final setting, \texttt{motion + MCDM (ReID+Flow)}, incorporates both the re-identification and flow losses.
For synthetic videos with \(64\) frames, we achieves competitive FID (\(47.1\)), the best FVD$_{16}$ (\(225.1\)) and IS (\(2.19\)), outperforming the motion-only baseline (FID: \(48.8\), FVD$_{16}$: \(244.2\), IS: \(2.15\)).
For synthetic videos with \(128\) frames, MCDM also improves temporal coherence, reducing the FVD$_{16}$ to \(262.8\) and FVD$_{128}$ to \(553.2\).
These results confirm that combining both pseudo-supervision signals yields the most effective motion representations and the strongest synthesis performance. While our label-free MCDM does not surpass EF-conditioned methods that rely on expert annotations, it provides a compelling alternative that eliminates the need for manual labels, offering a scalable solution for clinical ultrasound applications.

\begin{figure}[tb]
    \centering
    \resizebox{0.99\columnwidth}{!}{%
        \begin{tabular}{ccc}
            \begin{subfigure}[b]{0.2\linewidth}
                \centering
                \includegraphics[width=\linewidth]{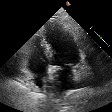}
                \caption{Real~\cite{ouyang2020video-based}}
            \end{subfigure}
            \begin{subfigure}[b]{0.2\linewidth}
                \centering
                \includegraphics[width=\linewidth]{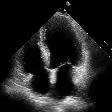}
                \caption{Baseline 64}
            \end{subfigure}
            \begin{subfigure}[b]{0.2\linewidth}
                \centering
                \includegraphics[width=\linewidth]{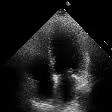}
                \caption{Baseline 128}
            \end{subfigure}
            \begin{subfigure}[b]{0.2\linewidth}
                \centering
                \includegraphics[width=\linewidth]{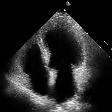}
                \caption{Baseline 192}
            \end{subfigure}
            \begin{subfigure}[b]{0.2\linewidth}
                \centering
                \includegraphics[width=\linewidth]{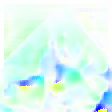}
                \caption{Optical flow}
            \end{subfigure}
            \\
            \begin{subfigure}{0.2\linewidth}
                \centering
                \includegraphics[width=\linewidth]{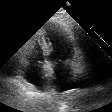}
                \caption{EchoSyn~\cite{reynaud2024echonet}}
            \end{subfigure}
            \begin{subfigure}[b]{0.2\linewidth}
                \centering
                \includegraphics[width=\linewidth]{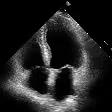}
                \caption{MCDM 64}
            \end{subfigure}
            \begin{subfigure}[b]{0.2\linewidth}
                \centering
                \includegraphics[width=\linewidth]{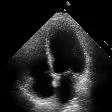}
                \caption{MCDM 128}
            \end{subfigure}
            \begin{subfigure}[b]{0.2\linewidth}
                \centering
                \includegraphics[width=\linewidth]{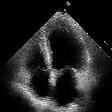}
                \caption{MCDM 192}
            \end{subfigure}
            \begin{subfigure}[b]{0.2\linewidth}
                \centering
                \includegraphics[width=\linewidth]{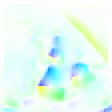}
                \caption{Optical flow}
            \end{subfigure}
            
        \end{tabular}%
    }
    \caption{Qualitative comparison of frames generated by different diffusion models. }
    \label{fig:qualitativevideo}
\end{figure}

\noindent\textbf{Qualitative comparison. } Fig.~\ref{fig:qualitativevideo} presents a qualitative comparison of frames generated by different diffusion models. For reference, Fig.~\ref{fig:qualitativevideo} (a) shows a real frame from the EchoNet-Dynamic dataset, while Fig.~\ref{fig:qualitativevideo} (f) shows a synthetic frame produced by EchoSyn~\cite{reynaud2024echonet}. Fig.~\ref{fig:qualitativevideo} (b–d) show sample videos with $64$, $128$, and $192$ frames, generated by diffusion models conditioned on motion features extracted using the baseline feature extractor.
In contrast, Fig.~\ref{fig:qualitativevideo} (g–i) show videos generated with our proposed MCDM, conditioned on motion features extracted by MAFE. The generated frames exhibit clearer cardiac structures and stronger spatial coherence, consistent with the improvements observed in FID and FVD, highlighting the effectiveness of our self-supervised framework without reliance on labelled supervision. Fig.~\ref{fig:qualitativevideo} (e) and (j) show examples of optical flow generated by UnSAMFlow~\cite{yuan2024unsamflow} with MedSAM~\cite{ma2024segment}, illustrating that our approach effectively captures cardiac motion.

\noindent\textbf{Ablation study. } Table~\ref{tab:ablationlamda} analyses the effect of the hyperparameters \(\lambda_1\) and \(\lambda_2\), which weight the re-identification and flow losses, respectively. These parameters control the relative influence of the pseudo appearance and pseudo flow supervision during feature extractor training. The best performance is achieved with \(\lambda_1=1\) and \(\lambda_2=0.01\), yielding the highest PSNR (23.31).

\begin{table}[tb]
  \caption{PSNR results for different hyperparameters  of \(\lambda_1\)
 (re-identification loss weight) and \(\lambda_2\) (flow loss weight).}
  \label{tab:ablationlamda}
  \begin{center}
  \begin{tabular*}{0.9\textwidth}{@{\extracolsep{\fill}} l c c c c c @{}}
    \toprule
     & \(\lambda_2 =0\) & \(\lambda_2 =0.005\) & \(\lambda_2 =0.01\) & \(\lambda_2 =0.05\) & \(\lambda_2 =0.1\) \\
    \midrule
   \(\lambda_1 =0\) & 23.17 & 23.24 & 23.21  & 23.20 & 23.13 \\
    \(\lambda_1 =1\) & 23.00 & 23.00 & \textbf{23.31} & 23.19 & 23.14 \\
    \(\lambda_1 =5\) & 23.24 & 23.23 & 23.29 & 23.19 & 23.15 \\
    \(\lambda_1 =10\) & 23.22 & 23.23 & 22.96 & 23.15 & 23.16 \\
    \bottomrule
  \end{tabular*}
  \end{center}
  \vspace{-3mm}
\end{table}

\section{Conclusion}

We propose a self-supervised Motion Conditioned Diffusion Model (MCDM) for realistic echocardiography video synthesis. 
The framework incorporates a Motion and Appearance Feature Extractor (MAFE) that disentangles motion and appearance features from ultrasound videos.
We design re-identification and optical flow losses guided by pseudo ground truth to enhance feature learning. Unlike prior approaches, our latent diffusion model is conditioned on motion features, removing dependence on clinically annotated anatomical parameters. Experiments on the EchoNet-Dynamic dataset demonstrate that MCDM achieves high-quality, temporally coherent video generation. 
Although MCDM remains behind EF-conditioned models, this trade-off underscores the value of eliminating manual labels and points toward scalable, privacy-preserving generative modelling for future clinical deployment.
To the best of our knowledge, this is the first work to apply self-supervised conditioning in latent diffusion models for echocardiography, highlighting the potential of unlabelled data to advance clinical imaging applications.

\paragraph{Acknowledgments:} HPC resources were provided by the Erlangen National High Performance Computing Center (NHR@FAU), under the NHR projects b143dc and b180dc. NHR is funded by federal and Bavarian state authorities, and NHR@FAU hardware is partially funded by the DFG - 440719683. H.R. was supported by Ultromics Ltd., the UKRI Centre for Doctoral Training in Artificial Intelligence for Healthcare  (EP / S023283/1). We acknowledge the use of Isambard-AI National AI Research Resource (AIRR)~\cite{mcintosh2024isambard}. Isambard-AI is operated by the University of Bristol and is funded by the UK Government’s DSIT via UKRI; and the Science and Technology Facilities Council [ST/AIRR/I-A-I/1023]. The authors received funding from the ERC-project MIA-NORMAL 101083647,  DFG 513220538, 512819079, and by the state of Bavaria (HTA).

\bibliographystyle{./styles/bibtex/splncs03}
\bibliography{main}

% \begin{thebibliography}{6}

% \end{thebibliography}
\end{document}